\documentclass[10pt,twocolumn,letterpaper]{article}

\usepackage{iccv}
\usepackage{times}
\usepackage{epsfig}
\usepackage{graphicx}
\usepackage{amsmath}
\usepackage{amssymb}

\usepackage{algorithm}
\usepackage{algpseudocode}
\usepackage[table]{xcolor}

\usepackage[accsupp]{axessibility}  % Improves PDF readability for those with disabilities.

\usepackage{subcaption}
\usepackage[pagebackref=true,breaklinks=true,letterpaper=true,colorlinks,bookmarks=false]{hyperref}

\iccvfinalcopy % *** Uncomment this line for the final submission

\def\iccvPaperID{8945} % *** Enter the ICCV Paper ID here

\ificcvfinal\fi

\begin{document}

%%%%%%%%% TITLE
\title{DINAR: Diffusion Inpainting of Neural Textures for One-Shot Human Avatars}

\author{David Svitov\\
Samsung Research\\
{\tt\small d.svitov@samsung.com}
\and
Dmitrii Gudkov\\
Samsung Research\\
{\tt\small d.gudkov@samsung.com}
\and
Renat Bashirov\\
Samsung Research\\
{\tt\small r.bashirov@samsung.com}
\and
Victor Lempitsky\\
Cinemersive Labs\\
{\tt\small victor@cinemersivelabs.com}
\vspace{-0.5em}
}
\maketitle
\thispagestyle{plain}
\pagestyle{plain}
\ificcvfinal\thispagestyle{empty}\fi

%%%%%%%%% ABSTRACT
\begin{abstract}
   We present DINAR, an approach for creating realistic rigged fullbody avatars from single RGB images. Similarly to previous works, our method uses neural textures combined with the SMPL-X body model to achieve photo-realistic quality of avatars while keeping them easy to animate and fast to infer. To restore the texture, we use a latent diffusion model and show how such model can be trained in the neural texture space. The use of the diffusion model allows us to realistically reconstruct large unseen regions such as the back of a person given the frontal view. The models in our pipeline are trained using 2D images and videos only. In the experiments, our approach achieves state-of-the-art rendering quality and good generalization to new poses and viewpoints. In particular, the approach improves state-of-the-art on the SnapshotPeople public benchmark.
\end{abstract}

%%%%%%%%% BODY TEXT
\section{Introduction}

The use of fullbody avatars in the virtual and augmented reality applications~\cite{mystakidis2022metaverse} is one of the drivers behind the recent surge of interest in fullbody photorealistic avatars~\cite{remelli2022drivable, alldieck2022photorealistic, prokudin2021smplpix}. Apart from the realism and fidelity of avatars, the ease of acquisition of new personalized avatars is of paramount importance. Towards this end, several works propose methods to restore 3D textured model of a human from a single image~\cite{saito2019pifu, saito2020pifuhd, he2020geo, alldieck2022photorealistic} but such models require additional efforts to produce rigging for animation. The use of additional rigging methods significantly complicates the process of obtaining an avatar and often restricts the poses that can be handled. At the same time, some of the recent methods use textured parametric models of human body~\cite{xu20213d, lazova2019360} while applying inpainting in the texture space. Current texture-based methods, however, lack photo-realism and rendering quality.

An alternative to using classical RGB textures directly is to use deferred neural rendering~\cite{thies2019deferred}. Such approaches make it possible to create human avatars controlled by the parametric model \cite{loper2015smpl, prokudin2021smplpix}. The resulting avatars are more photo-realistic and easier to animate. However, existing approaches require a video sequence to create an avatar \cite{raj2021anr}. The StylePeople system~\cite{grigorev2021stylepeople}, which is also based on deferred neural rendering and parametric model, provides an opportunity to create avatars from single images, however the quality of rendering for unobserved body parts is low. 

\begin{figure*}[t]
  \centering
  \includegraphics[width=0.95\linewidth]{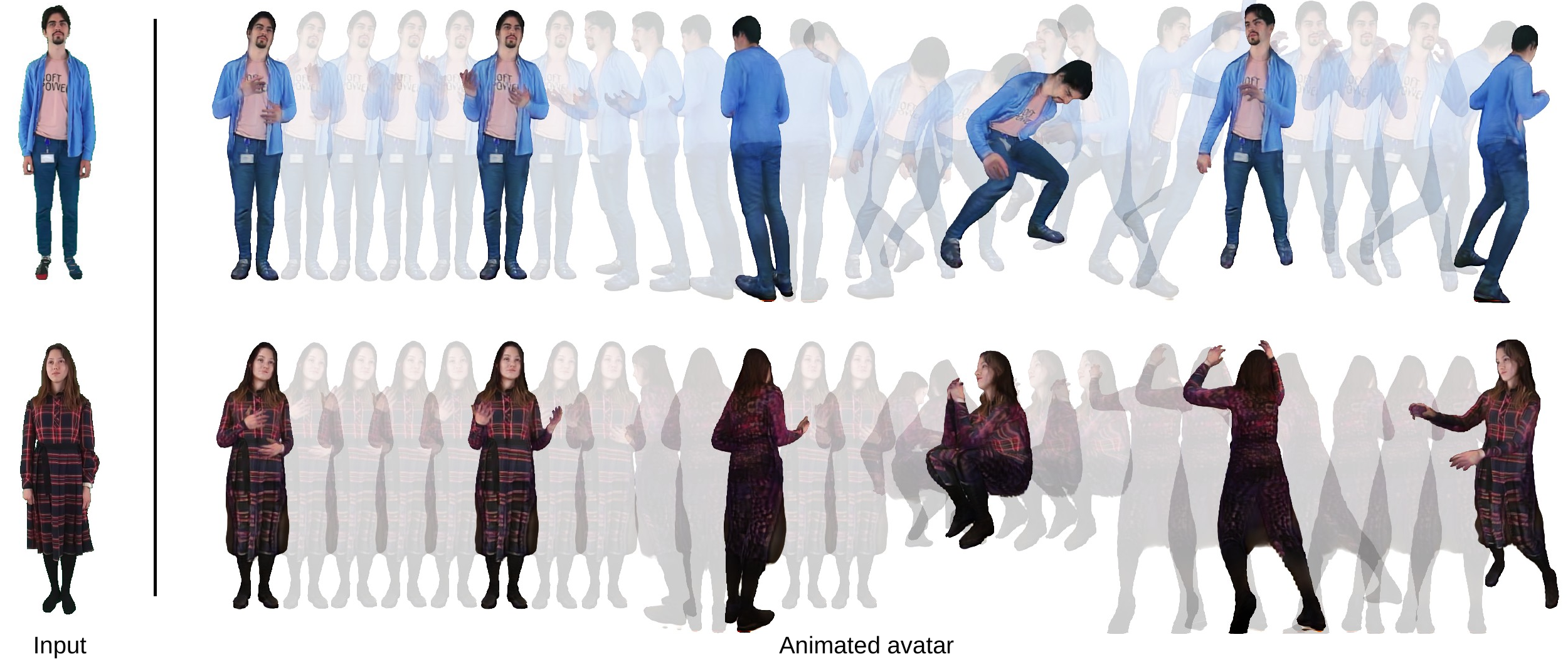}
   \caption{\textbf{Animations of one-shot avatars.} We generate avatars of previously unseen people from single images and animate the avatars by changing their SMPL-X poses. Our model produces plausible results for new poses and views, even for complex and loose garments like skirts and can handle complex poses (such as hands near the body) gracefully.}
   \label{fig:Azure}
\end{figure*}

We propose a new method to create photo-realistic animatable human avatars from a single photo. To make the avatars animatable we leverage neural texture approach~\cite{thies2019deferred} along with the SMPL-X parametric body model~\cite{prokudin2021smplpix}. We propose a new architecture for generating the neural textures, in which the texture comprises both the RGB part explicitly extracted from the input photograph by warping and additional neural channels obtained by mapping the image to a latent vector space and decoding the result into the texture space. As is customary with neural rendering~\cite{thies2019deferred, grigorev2021stylepeople, raj2021anr}, the texture generation is trained in an end-to-end fashion with the rendering network.

To restore the neural texture for unobserved parts of the human body we develop a diffusion model~\cite{ho2020denoising}. This approach allows us to obtain photo-realistic human avatars from single images. In the presence of multiple images, we can merge neural textures corresponding to different images while restoring parts that are still missing by diffusion-based inpainting. The use of diffusion for inpainting distinguishes our approach from several previous works~\cite{lazova2019360, he2021arch++, grigorev2019coordinate} including StylePeople~\cite{grigorev2021stylepeople} that rely on generative adversarial framework~\cite{NIPS2014_5ca3e9b1} to perform inpainting of human body textures. As in other image domains~\cite{dhariwal2021diffusion, saharia2022image}, we found that the use of diffusion alleviates problems with mode collapse and allows to obtain plausible samples from complex multi-modal distributions. To the best of our knowledge, we are the first to extend the diffusion models to the task of generating human body textures.

To sum up, our contributions are as follows:
\begin{itemize}
    \item We propose a new approach for modeling human avatars based on neural textures that combine the RGB and the latent components.   
    \item We adapt the diffusion framework for neural textures and demonstrate that it is capable of inpainting such textures.
    \item We demonstrate the ability of our system to build realistic animatable avatars from a single photograph.
\end{itemize}

%<
The proposed approach allows us to obtain a photorealistic animatable person's avatar from a single image. Specifically, when the input photograph is taken from the front, we observe that the person's back is restored in a consistent photo-realistic manner.
%>
We demonstrate the effectiveness and accuracy of our approach on real-world images from SnapshotPeople \cite{alldieck2018video} public benchmark and images of people in natural poses.

%-------------------------------------------------------------------------
\section{Related work}

\begin{figure*}
  \centering
  \includegraphics[width=1.0\linewidth]{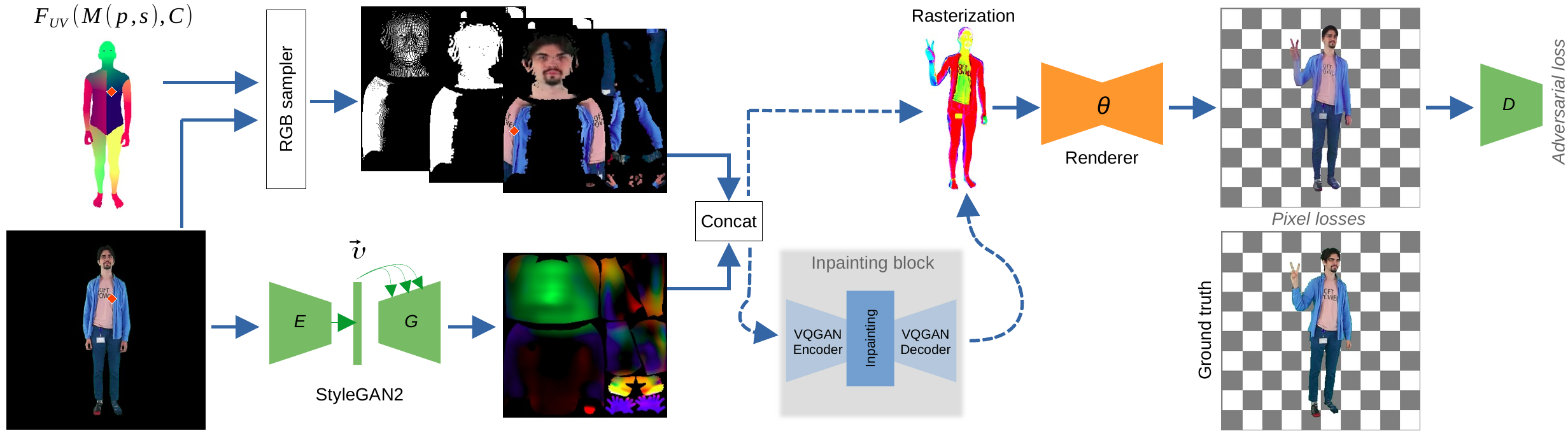}
   \caption{\textbf{Overview of our neural texture avatar pipeline:} Given an RGB image and the corresponding SMPL-X model as input, we use a UV-map to sample the RGB texture. Using an encoder and a StyleGAN2 generator, we convert the input image into a neural texture. We rasterize SMPL-X with concatenation of the neural and RGB textures and translate it to the RGB+mask output using the rendering network. The inpainting block replaces missing texture fragments and is described in detail in Section \ref{sec:Inpainting}.}
   \label{fig:VAE}
\end{figure*}

\textbf{Full body human avatar.}
Many avatar systems are based on parametric human body models, of which the most popular are the SMPL~\cite{loper2015smpl} body model as well as the SMPL-X model~\cite{prokudin2021smplpix} which augments SMPL with facial expressions and hand articulations. Such models represent human body without garments or hair. Approaches based on deferred neural rendering (DNR)~\cite{thies2019deferred} or neural radiance fields (NeRF)~\cite{mildenhall2021nerf} can be used to add clothing and perform photo-realistic rendering. DNR uses a multi-channel trainable neural texture and a rendering convolutional network to render the resulting avatars in a realistic way~\cite{raj2021anr, grigorev2021stylepeople}. It makes it easier to animate the resulting avatar. NeRF uses sampling along a ray in implicit space for rendering and allows one to extract accurate and consistent geometry~\cite{saito2019pifu, huang2020arch, alldieck2022photorealistic}.

\textbf{Video full body avatar.}
The parametric SMPL model is used to obtain a photo-realistic human avatar from a monocular video \cite{alldieck2018detailed, alldieck2018video, weng2022humannerf, raj2021anr}. The use of a video enables the creation of high-quality avatars due to the large amount of input information. However, the applicability of such approaches is limited by the difficulty of obtaining such videos.

A number of techniques \cite{raj2021anr, grigorev2021stylepeople, larionov2023morf} use neural textures to simulate out-of-mesh details in order to create avatars from monocular video. For a more temporally consistent representation of loose clothing, some methods \cite{zhao2022high, alldieck2018detailed, alldieck2018video} use vertex displacements in addition to the texture.
The highest photorealism and temporal consistency in the video avatar task are currently achieved by NeRF-based methods \cite{peng2021animatable, weng2022humannerf, jiang2022neuman}. Such methods use SMPL to translate an avatar to the implicit canonical space. Then render it by integrating along rays from a camera. 

\textbf{One-shot full body avatar.} 
One-shot avatar approaches reconstruct human body avatars from a single image. Early works achieved this by inpainting partial RGB textures~\cite{xu20213d, lazova2019360}. These approaches did not allow realistic modeling of avatars with clothing. More recent works on one-shot body modeling relied on implicit geometry and radiance models, which predict occupancy and color with the multi-layer perceptron conditioned on feature vectors extracted from the input image~\cite{saito2019pifu, alldieck2022photorealistic,he2020geo}. While this line of work often manages to recover intricate geometric details, the realism of the recovered texture in the unobserved parts is usually limited.

The ARCH system~\cite{huang2020arch} uses the rigged canonical space to build avatars suitable for animations. ARCH++~\cite{he2021arch++} improves the quality of resulting avatars by revisiting the major steps of ARCH. They also solve the challenge of unseen surfaces by synthesizing the back views of a person from the front view. PHORHUM~\cite{alldieck2022photorealistic} improves the modeling of geometry by adding the reasoning about scene illumination and albedo of the surface.
S3F \cite{corona2023structured} also enables changing the avatar lighting. But it takes advantage of 3D Features connected to the vertices of the parametric model, which allows one to animate the resulting avatar.
ICON \cite{xiu2022icon} uses local features to avoid the dependence of the reconstructed geometry on the global pose. Their approach first estimates separate models from each view and then merges the models using SCANimate \cite{saito2021scanimate}. The method uses RGB textures applied to reconstructed geometry, which limits the rendering photo-realism. 

An alternative approach to getting one-shot avatars is to use avatar generative models as proposed in StylePeople \cite{grigorev2021stylepeople}. The authors circumvent the need to reconstruct the unseen parts by exploiting the entanglement in the GANs latent space. Unfortunately, the imperfection of their underlying generative model often leads to implausible appearance of unobserved parts.

\textbf{Diffusion models.} Diffusion models \cite{sohl2015deep} are probabilistic models for learning the distribution of $p(x)$ by gradual denoising of a normally distributed variable. Such denoising corresponds to learning the inverse process for a fixed Markov chain of length $\mathrm{T}$. The most successful models for image generation \cite{dhariwal2021diffusion, ho2020denoising, rombach2022high} use the reweighted variant of the variational lower bound on $p(x)$. These models can also be interpreted as denoising autoencoders $\epsilon_\omega(x_t, t); t=1, ..., \mathrm{T}$ with shared weights. These autoencoders learn to predict $x_{t-1}$ with reduced noise level over $x_t$. In \cite{ho2020denoising} was demonstrated that such denoising models can be trained with a simplified loss function:
\begin{equation} \label{eq:dm_loss}
L_{DM} = \mathbb{E}_{x, \epsilon \sim \mathcal{N}(0, 1), t}[||\epsilon - \epsilon_\omega(x_t, t)||_2^2],
\end{equation}
where $t$ is uniformly sampled from $\{1, ..., \mathrm{T}\}$. Our work builds on recent advances in inpainting with diffusion models \cite{saharia2022palette, lugmayr2022repaint, romero2022ntire}. In particular, we use latent diffusion \cite{rombach2022high} in our approach, which has been shown to be effective in inpainting of RGB images.

%<
%Diffusion models \cite{sohl2015deep} emerged as an alternative to generative models \cite{NIPS2014_5ca3e9b1}. In the human modeling domain, the diffusion models were shown to work very well for the task of human motion generation~\cite{tevet2022human, zhang2022motiondiffuse, ren2022diffusion}. RODIN \cite{Wang2022RodinAG} employs a diffusion framework to generate non-rigged head avatars as neural radiance fields represented by 2D feature maps. TEXTure \cite{Richardson2023TEXTureTT} uses text guidance and a pre-trained diffusion model to produce a view-consistent RGB texture of a given geometry. To the best of our knowledge, diffusion models have not yet been used to generate neural textures for 3D objects.
%>

%-------------------------------------------------------------------------

\section{Method}

Our approach has two components: the avatar generation model and the inpainting model. The scheme of our avatar generation model is presented in Fig. \ref{fig:VAE}. The model reconstructs the neural texture from the input image using two pathways and then uses the texturing operation as well as the neural rendering to synthesize images of the avatar. Then we train the inpainting model based on the denoising diffusion probabilistic model (DDPM) \cite{ho2020denoising} on top of the pretrained avatar generation model. This model is used to restore body areas unpresented in the input image.

\subsection{Avatar generation model}

Our approach creates a 3D rigged avatars of clothed human using several neural networks and the texturing modules trained together in the end-to-end fashion. The overall architecture takes as input an RGB image and a parametric \mbox{SMPL-X}~\cite{SMPL-X:2019} body model. During training, we fit \mbox{SMPL-X} models to the train images using an implementation of \mbox{SMPLifyX}~\cite{SMPL-X:2019} approach with an additional segmentation loss that allows us to match human silhouettes better. 

%In more detail, we use an SMPL-X fixed-topology mesh $M(p, s)$, driven by sets of pose parameters $p$ and shape parameters $s$. We also use the UV-map function $F_\textrm{UV}(M, C_\textrm{target})$ for the texture mapping depending on a mesh and target camera position $C_\textrm{target}$. For SMPL-X, we employ a customized UV unwrapping with a front cut to avoid difficult to inpaint back-view seams. The pose parameters $p_\textrm{target}$ are used to rig the mesh. The texturing function $F_\textrm{UV}$ generates a UV-map of size $H \times W \times 2$, where $H$ and $W$ determine the size of the output image, and for each pixel the texture coordinates $[i, j]$ on the $L$-channeled texture $T$ are specified. We thus use $F_\textrm{UV}$ in the rasterizer $R(F_\textrm{UV}, T)$ to map the pixels in the output image to the features of the texels of the neural texture $T$. The rasterizer $R$ thus produces the image of size $H \times W \times L$. 

In more detail, we use an SMPL-X fixed-topology mesh $M(p, s)$, driven by sets of pose parameters $p$ and shape parameters $s$. Here the pose parameters $p$ are used to rig the mesh. Information about person appearance is stored in the $L$-channeled neural texture $T$. One-to-one correspondence between the mesh $M$ and the texture $T$ is set as a UV-map. For SMPL-X, we employ a customized UV-map with a front cut to avoid difficult to inpaint back-view seams. For the mapping of texture $T$ to the output image we use the resterizer $R(T, M, C_\textrm{target})$ depending on the mesh $M$ and target camera position $C_\textrm{target}$. As result, the rasterizer $R$ produces an image of size $H \times W \times L$. Where $H$ and $W$ determine the output image size and $L$ is the texture features size. 

We set parameters of the rasterizer $R$ so that $H$ and $W$ correspond to the height and width of the input RGB image $I_\textrm{rgb}$. In this case, we can use UV not only to map feature vectors from the neural texture $T$, but also to sample color values from the input image $I_\textrm{rgb}$ into the texture space: $T_\textrm{rgb}$. We transfer the color value from $I_\textrm{rgb}$ to the texture $T_\textrm{rgb}$ point specified by the UV-map. This RGB texture allows us to explicitly save information about high-frequency details and original colors, which are hard to preserve when mapping the whole image to a vector of limited dimensionality (as discussed below). We additionally apply inpainting of small gaps with averaging neighbor pixels to fill the gaps in $T_\textrm{rgb}$. We also save the binary map of sampled pixels to the $B_\textrm{smp}$ and the map of the sampled and inpainted pixels to the $B_\textrm{fill}$.

The main part of the neural texture is $T_\textrm{gen}$. It has the number of channels $L=16$ and is generated using the encoder-generator architecture $T_\textrm{gen} = G(E(I_\textrm{rgb}))$. The encoder $E$ is based on the StyleGAN2 \cite{karras2020analyzing} discriminator architecture and compresses the input image $I_\textrm{rgb}$ to a vector of dimension 512. The generator $G$ has the architecture of the StyleGAN2 generator and converts the vector into a $T_\textrm{gen}$ neural texture with the number of channels $L=16$ as in StylePeople \cite{grigorev2021stylepeople}.

The final neural texture used in our method has a dimension of $256 \times 256 \times 21$ and consists of the concatenation of: the generated texture $T_\textrm{gen}$ ($256 \times 256 \times 16$), the texture $T_\textrm{rgb}$ sampled from the RGB image ($256 \times 256 \times 3$) and the two binary segmentation maps ($B_\textrm{smp}$ and $B_\textrm{fill}$):
\begin{equation} \label{eq:Texture}
T = T_\textrm{gen} \oplus T_\textrm{rgb} \oplus B_\textrm{smp} \oplus B_\textrm{fill}.
\end{equation}
We note that such an approach with the explicit use of RGB channels as part of the neural texture was originally used in \cite{thies2019deferred}.

We use the neural renderer $\theta(R(T, M, C_\textrm{target}))$ to translate the rasterized image with $L+3+1+1$ channels into $I_\textrm{rend}$ output RGB image. Neural renderer $\theta$ has a U-Net \cite{ronneberger2015u} architecture with ResNet \cite{he2016deep} blocks. We train the renderer $\theta$ jointly with the neural texture encoder $E$ and generator $G$. Thus, rendering an avatar with a texture $T$ (Eq. \ref{eq:Texture}) in a new pose $p_\textrm{target}$ has the following form:

\begin{equation} \label{eq:rendering}
I_\textrm{rend} = \theta(R(T, M(p_\textrm{target}, s), C_\textrm{target}))
\end{equation}

During training, we minimize the following losses: the difference loss $L_2$ and the perceptual loss $LPIPS$ \cite{zhang2018perceptual} between the rendered $I_\textrm{rend}$ and ground truth $I_\textrm{GT}$ images; the Dice loss \cite{sudre2017generalised} between the ground truth and the predicted segmentation masks. We calculate $L2$ and $LPIPS$ for the entire image and additionally with a weight of $0.1$ for the area with a face, since it has been demonstrated in~\cite{fruhstuck2022insetgan} that the face is very important for human perception. Additionally, we use an adversarial loss to make $I_\textrm{rend}$ look more realistic and sharp. We use nonsaturating adversarial loss $Adv$~\cite{NIPS2014_5ca3e9b1} with the StyleGAN2 discriminator $D$ with R1-regularization \cite{mescheder2018training}. The overall loss thus has the following form: %\vspace{-1em}

\begin{equation} \label{eq:losses}
\begin{split}
Loss = & \lambda_1 \cdot L_2 + \lambda_2 \cdot LPIPS + \lambda_3 \cdot Dice\\
       & + \lambda_4 \cdot Adv + \lambda_5 \cdot R1_\textrm{reg}.
\end{split}
\end{equation}

We describe the choice of hyperparameters $\lambda_{1..5}$ in Section \ref{sec:Experiments}.

Our training is performed on multi-view image sets such as sets of frames of the video (or such as sets of renders of 3D models). The training data and network training details are discussed in Section \ref{sec:Experiments}. During training, in each case we take two different frames from the same set, one as the input image $I_\textrm{rgb}$ and the other as the target image $I_\textrm{GT}$ (the two thus having different camera parameters as well as two different body pose parameters). This is essential for the pipeline to generalize to new camera positions $C$ and poses $p$. To accomplish that, the renderer and the texture generation modules learn to inpaint small fragments unseen in $I_\textrm{rgb}$.

\subsection{Texture merging} \label{sssec:merging}

%<
While the texture generator and the renderer learn to compensate for the small amount of the unseen surfaces that may be present in the target view, we have found that such ability is limited to small changes in body pose and/or camera parameters. 
%>

\begin{figure}
  \centering
  \includegraphics[width=1.0\linewidth]{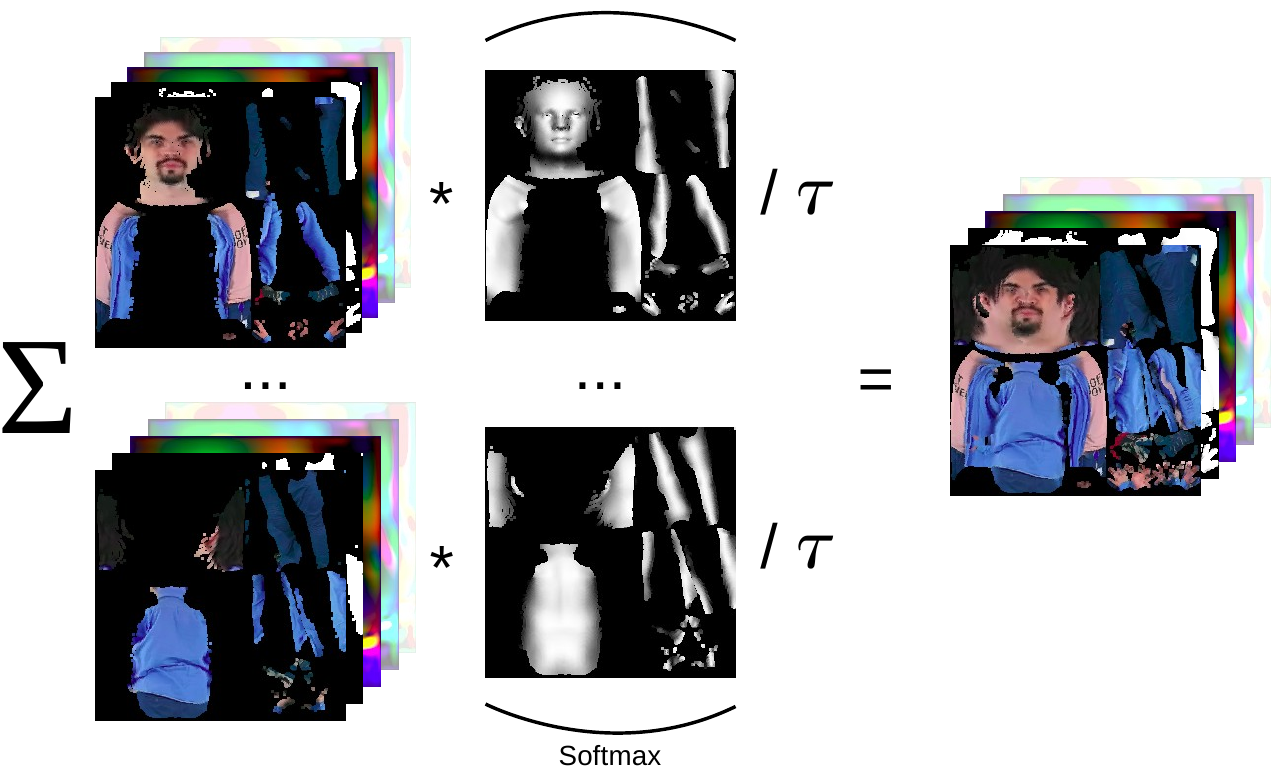}

   \caption{\textbf{Texture merging:} We get the full neural texture as the weighted sum of textures from different view points. As weights, we use the angles between the normal vectors and the direction of the camera. For simplicity, only the RGB channels are shown in the figure, but the merging affects all channels.}
   \label{fig:Merge}
   %\vspace{-1.5em}
\end{figure}

The easiest way to obtain an avatar that can be rendered from arbitrary angles is to create it from several images by merging the corresponding neural textures. For that we can use a simple blending scheme (Fig. \ref{fig:Merge}). Specifically, assume the $N$ input images $I_\textrm{rgb}^i$ from different view points are given that result in $N$ neural textures $T^i$. These textures naturally cover different areas of the body visible in different input images. To combine these textures, we define the function $F(T^1 ... T^N, \lambda^1 ... \lambda^N)$ shown in Fig. \ref{fig:Merge}. The $\lambda^i$ is auxiliary information about the visible in $I_\textrm{rgb}^i$ part of the texture.

As auxiliary information $\lambda^i$ we use the angles between the normal vectors of corresponding point of the mesh $M^i$ and direction vector of the camera. Thus $\lambda^i$ defines how frontal each texture point is to the camera. We perform texture merging utilizing this information, emphasizing the more frontal pixels for each $T^i$. We aggregate textures $T^i$ using  weighted average with weights calculated as $\vec{w} = softmax(\frac{\vec{\lambda}}{\tau})$. The $\tau$ factor controls the sharpness of edges at the junction of merged textures. The final texture is calculated as:

\begin{equation}
T = F(T^1 ... T^N, \lambda^1 ... \lambda^N) = \sum_{i=1}^{N} {T^i \cdot w^i}.
\end{equation}

This technique allows us to get few-shot avatars by merging one-shot avatars for different views. More sophisticated blending schemes such as pyramid blending~\cite{ogden1985pyramid} or Poisson blending~\cite{perez2003poisson} can be also used.

\begin{figure}
  \centering
  \includegraphics[width=1.0\linewidth]{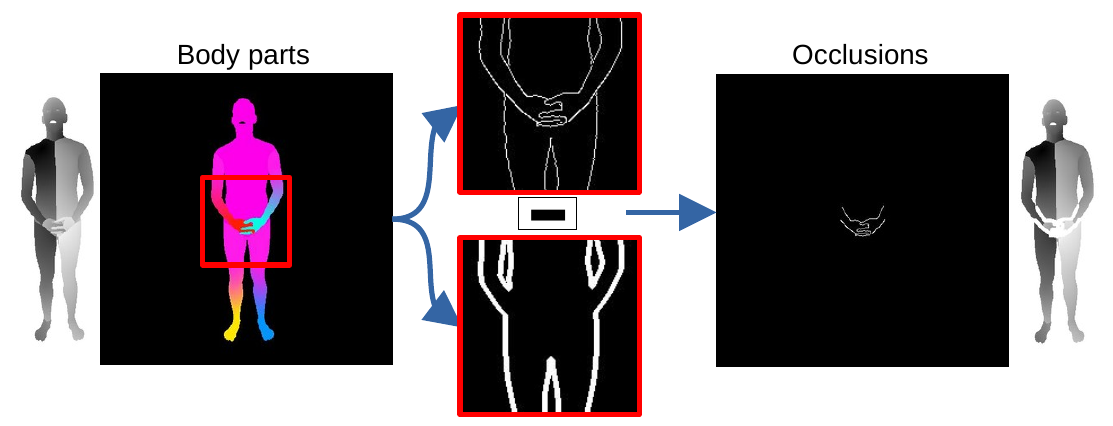}

   \caption{\textbf{Occlusions detection.} We use rasterization with the colored body parts texture to detect areas occluded by limbs. The detected areas are masked out of the final UV-render to reduce inaccuracies in RGB texture sampling.}
   \label{fig:Occlusions_small}
   %\vspace{-1.5em}
\end{figure}

\begin{figure*}
  \centering
  \includegraphics[width=1.0\linewidth]{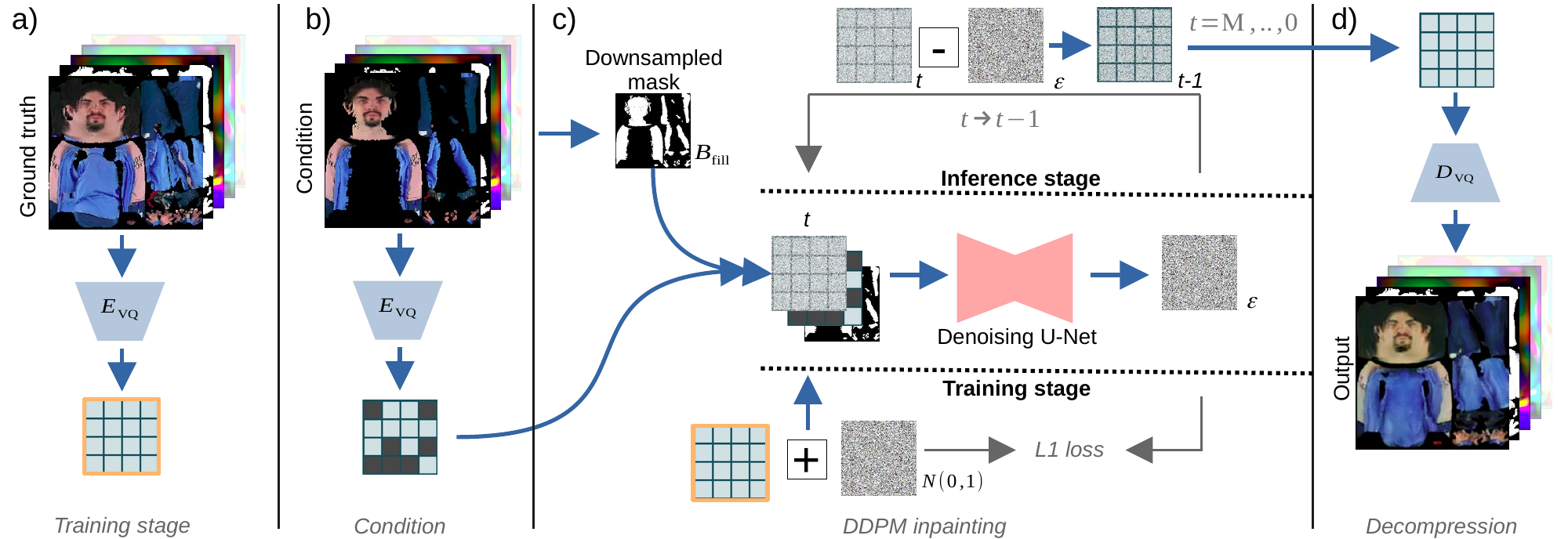}

   \caption{\textbf{Texture inpainting:} a) In the training step, we reduce the size of the ground truth and partial (b) neural textures using the VQGAN encoder. c) At train U-Net learns to remove noise from Ground truth texture for a given step $t$. At inference, we iteratively apply $M$ steps of denoising to sample the image. d) The sampled texture is transformed to the original size by the VQGAN decoder.}
   \label{fig:Inpaint}
   \vspace{-1.0em}
\end{figure*}

\subsection{The inpainting model}\label{sec:Inpainting}

As the final piece of our approach, we train a network that can create complete neural textures from a single photographs. We do that using supervised learning, where we use the incomplete textures based on single photographs as inputs and the combined textures aggregated from multiple views as ground truth.

Since the distribution of plausible (``correct'') complete textures given the input partial texture is usually highly complex and multi-modal, we use the denoising diffusion probabilistic model (DDPM) framework~\cite{ho2020denoising} and train a denoising network instead of the direct mapping from the input to the output.

\textbf{Reducing texture space.} As described above, in our experiments the neural texture $T$ has a resolution of $256 \times 256 \times 21$. This leads to a huge memory requirements during the diffusion model training. In order to reduce memory consumption and improve the network convergence, we first reduce the neural texture size using VQGAN autoencoder~\cite{esser2020taming} analogously to the reduction of an RGB image size performed in \cite{rombach2022high}. As demonstrated in Fig.~\ref{fig:VAE} VQGAN is added as an alternative branch for the input of the renderer $\theta$. After the pretraining of VQGAN, the pipeline is finetuned end-to-end in order to adapt the renderer to VQGAN decompression artifacts in the neural texture $T_\textrm{res}$.

We use several loss functions to train the VQGAN autoencoder to more accurately restore neural textures. To improve the visual quality of the avatar after texture decompression, we use the loss functions in the RGB space. We render the avatar as described in (\ref{eq:rendering}) with the restored texture $T_\textrm{res}$ and optimize the loss function (\ref{eq:losses}). Also, we use an additional L2 loss in texture space $||T - T_\textrm{res}||_2^2$ for additional regularization and preservation of neural texture properties for all views.

\textbf{Texture inpainting.} After adding the VQGAN branch to the pipeline, we train the DDPM model in its latent space. Thus, the diffusion model is applied to $T_\textrm{c}^j = E_\textrm{VQ}(T^j)$ with size $64 \times 64 \times 3$, obtained after the compression of the single-view based texture $T^j$ by the VQGAN encoder $E_\textrm{VQ}$. Index $j$ in our experiments corresponds to the front view. Following \cite{rombach2022high} we train DDPM using a U-Net architecture with attention. We condition the denoising model with $T_\textrm{c}^j \oplus b(B_\textrm{fill}^j)$, where $b$ is a bilinear resize to the spatial size of $T_\textrm{c}^j$. As the input to the model, we feed in the concatenation of the condition and the compressed merged texture $T_c$ corrupted with normally-distributed noise corresponding to the diffusion timestep $t$. The U-Net architecture thus trains to denoise the input $T_\textrm{c}$ by minimizing the loss (\ref{eq:dm_loss}). This allows $T_\textrm{c}$ to be iteratively derived from pure Gaussian noise during conditional inference.

As mentioned above, we train diffusion inpainting (Fig. \ref{fig:Inpaint}) using the merged textures (section \ref{sssec:merging}) as ground truth. Here, we generate textures from a dataset of 3D human scans. Multi-view dataset of people photographs with good angle coverage could be used as training data as well, and we chose to use the scan render dataset solely because of its availability to us. The training dataset is discussed in section \ref{sec:Experiments}.

\subsection{Inference} \label{sec:Inference}

We perform inference in two stages. To get an animated avatar from a photo, we first get a partial texture from the input image, and then we inpaint it with the DDPM model.

To create the partial texture $T^j$ we inference the model shown in Figure \ref{fig:VAE} with input image $I_\textrm{rgb}^j$. We also apply several techniques to improve avatar quality in the inference stage. To enhance the texture details in the visible part, we perform a few optimization steps of RGB channels with gradients from the differentiable renderer for the input image. To reduce the impact of \mbox{SMPL-X} fitting  imperfections, we detect areas of human self-occlusion in the input image (Section \ref{sec:Oclusion}) and do not sample the texture along the overlap outline. The restoration of these areas are left for the inpainting process described below.

During the inpainting stage the DDPM restores the whole texture $T_\textrm{c}$ from the noise conditioned on the partial texture. To do this, we perform $M$ denoising steps starting with pure noise as shown in Figure \ref{fig:Inpaint}. We then transform $T_c$ into the restored full-size texture $T_\textrm{res}$ using the VQGAN decoder $D_\textrm{VQ}$. To retain all the details from the input view, we then merge the restored texture with the input texture $T^j$ using the $B_\textrm{fill}^j$ mask:
\begin{equation}
T = T_\textrm{res} \cdot (1 - B_\textrm{fill}^j) + T^j \cdot B_\textrm{fill}^j
\end{equation}
The resulting texture has all the details visible in the $I_\textrm{rgb}^j$ image, while the parts of a person invisible on $I_\textrm{rgb}^j$ are restored by the DDPM in the VQGAN latent space. An avatar with a texture $T$ obtained in this way can be rendered from new view points and in new poses.

\subsection{Addressing proxy-geometry imperfections} \label{sec:Oclusion}

We have found that due to imperfect SMPL-X meshes, pixels are wrongly sampled from one body part to another at the self-occluded areas (\eg hands in front of the body). This results in implausible renderings (Fig. \ref{fig:Ablation} (b, c)). 

To address this issue, we do not sample the RGB texture along the outline of such overlapping body parts. We employ rasterization with a colormap as a texture to find overlapping regions (Fig. \ref{fig:Occlusions_small}). 
We assigned each limb in the colormap a separate color and made the transition between them smooth using a color gradient. This enables us to avoid having seams in the rasterization. We detect edges in the colormap rasterization with Canny algorithm~\cite{Canny1986ACA}. We then determine a person's contours employing binarized SMPL-X rasterization. By taking the contours out of edges, we obtain an occlusion map. We use the resulting occlusion map to mask out areas in the UV-render. This enables us to rely on inpainting in later pipeline steps rather than sampling pixels in overlapped areas. 

%-------------------------------------------------------------------------

\section{Experiments}\label{sec:Experiments}

\begin{figure*}
  \centering
  \includegraphics[width=1.0\linewidth]{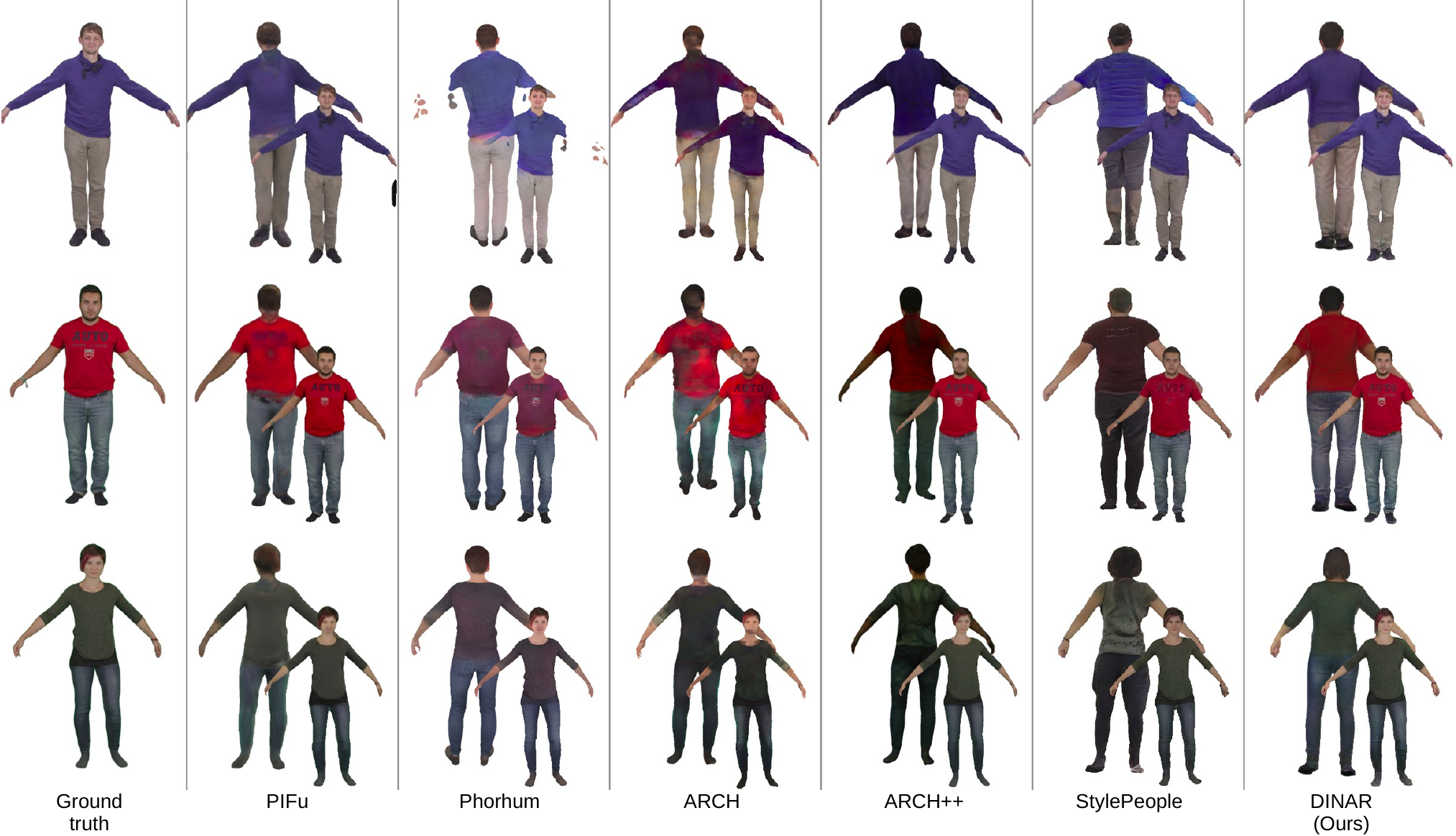}
   \caption{\textbf{Qualitative comparison on the SnapshotPeople benchmark.} We compare our method with state-of-the-art approaches for creating avatars: PIFu, PHORHUM, ARCH, ARCH++, StylePeople. For each method, we show a front view and a back view to evaluate the quality of back reconstruction. Results for PHORHUM, ARCH and ARCH++ provided by the authors and registered with ground truth images.}
   \label{fig:Snapshot}
   \vspace{-0.8em}
\end{figure*}

\subsection{Implementation details}

Now we present implementation details and hyperparameters values. Our model is trained on RGB images at the $512 \times 512$ resolution. First, we train the generator and the renderer with the next losses: the L2 loss, the LPIPS loss, the Dice loss~\cite{sudre2017generalised} and the adversarial nonsaturating loss~\cite{NIPS2014_5ca3e9b1} with R1 regularization~\cite{mescheder2018training}. We use a weighted sum of losses with the following weights. L2 loss with $\lambda_1 = 2.2$; LPIPS loss with $\lambda_2 = 1.0$; Dice loss with $\lambda_3 = 1.0$; Adversarial loss with $\lambda_4 = 0.01$. We use the lazy regularization R1 as proposed in~\cite{karras2020analyzing} every 16 iterations with the weight $\lambda_5 = 0.1$. To calculate LPIPS, we take a random $256 \times 256$ crop of the $512 \times 512$ images. We train the generator and the renderer end-to-end for 100,000 steps using the ADAM optimizer~\cite{kingma2014adam} with $2e{\text -}3$ learning rate and the batch size of four.

Then we train VQGAN to compress neural textures to $6 \times 64 \times 64$ tensors consisting of vectors of length six from a trainable dictionary with 8192 elements. We first train only the VQGAN branch for 300,000 steps. Then we finetune the pipeline end-to-end for additional 20,000 steps to reduce the renderer artifacts when processing neural textures after VQGAN. After that, we train the diffusion probabilistic model to restore missing parts of the texture. We use the U-Net architecture with the BigGAN~\cite{brock2018large} residual blocks for up- and downsampling and with attention layers on three levels of its feature hierarchy. To additionally prevent over-fitting, we use dropout with a probability of 0.5 in the residual blocks. We train the diffusion model for 50,000 iterations with AdamW optimizer~\cite{loshchilov2017decoupled} with a batch size of 128 and a learning rate of $1.0e{\text -}6$.

\subsection{Datasets}

\begin{table*}[]
\centering
\begin{tabular}{c|cccc|ccc}
                     & \multicolumn{4}{c|}{\textbf{Same view}} & \multicolumn{3}{c}{\textbf{Novel view}} \\
Method               & MS-SSIM $\uparrow$ & PSNR $\uparrow$ & LPIPS $\downarrow$ & DISTS $\downarrow$ & DISTS $\downarrow$ & ReID $\downarrow$\  & KID $\downarrow$ \\ \hline
PIFu                 & \cellcolor{green!15}0.9793  & \cellcolor{green!15}26.2828   & \cellcolor{green!15}0.0404  & \cellcolor{yellow!15}0.0706 &     0.1839       &      0.09769           &    0.0907          \\
Phorhum              &    0.9603                   &  24.2112                      &        0.0531               &    0.0948                   & \cellcolor{yellow!15}0.1564 &      0.09149           &   \cellcolor{yellow!15}0.0144    \\ \hline
ARCH                 &    0.9223                   &  20.6499                      &        0.0732               &    0.1432                   &     0.2039       &     0.09575            &     0.0974         \\
ARCH++               &    0.9526                   &  22.5729                      &        0.0540               &    0.0842                   &     0.1750       & \cellcolor{yellow!15}0.09098   &     0.0589         \\ \hline
StylePeople          &    0.9610                   &  23.0584                      &        0.0848               &    0.0852                   &     0.1698       &     0.13090            &     0.0530         \\
\textbf{DINAR (Ours)} & \cellcolor{yellow!15}0.9687 &  \cellcolor{yellow!15}24.4182 & \cellcolor{yellow!15}0.0504 & \cellcolor{green!15}0.0703  &  \cellcolor{green!15}0.1407    &  \cellcolor{green!15}0.07855  & \cellcolor{green!15}0.0133                          
\end{tabular}

\caption{\label{table:metrics} \textbf{Metrics comparison on the SnapshotPeople benchmark.} We compared our method not only with other parametric model based approaches (StylePeople), but also with approaches that restore geometry and require using additional methods for rigging (PIFu, Phorhum) or restore geometry in the canonical pose (ARCH, ARCH++).}
%\vspace{-1em}
\end{table*}

To train our pipeline we used only 2D images obtained by rendering Texel~\cite{Texel} dataset. We pretrained the neural texture generator and the neural renderer using 2D images of 13,000 people in diverse poses. We have noticed that diverse poses are crucial to train realistically animatable avatars. For each image, we obtained a segmentation mask using Graphonomy~\cite{Gong2019Graphonomy} 
%<
segmentation 
%>
and fit the SMPL-X parametric model using SMPLify-X~\cite{SMPL-X:2019}. We also used a segmentation Dice loss~\cite{sudre2017generalised} to improve the body shape of fitted SMPL-X.

To train VGQAN and the diffusion model, we used renders from Texel dataset. We acquired 3333 human scans from the Texel dataset. They are people in different clothing with various body shapes, skin tones and genders. We rendered each scan from 8 different views to get a multi-view dataset. We also augmented the renders with camera angle changes and color shifts. Thus, for each person in the dataset, we got 72 images. Note that any images from different views are suitable for training the model, not necessarily obtained from 3D scans.

We qualitively evaluate our avatars and their animations on AzurePeople~\cite{bashirov2021real} dataset (Fig. \ref{fig:Azure}). This dataset contains diverse dressed people standing in natural poses. We also quantitatively evaluate our pipeline on the SnapshotPeople \cite{alldieck2018video} public benchmark. It contains 24 videos of people rotating in A\nobreakdash-pose. We select frames with front and back views from each video to measure the accuracy of front and back reconstruction respectively. For each image we get a segmentation mask and SMPL-X fit as described above.

\begin{table}[t]
\centering
\begin{tabular}{m{8em}|ccc}
Method                     & MS-SSIM $\uparrow$ & PSNR $\uparrow$\  & LPIPS $\downarrow$ \\ \hline
RGB texture   & 0,8825       &     22,8041            &   0,1204          \\ \hline
Neural texture    & 0,8632       &     23,0935            &   0,1285    \\ \hline
Both textures   & 0,9126       &     24,1041            &   0,1005         \\ \hline
Both + Inpainting   & 0,9199       &     24,8510            &   0,0904         \\ \hline
Both + Inpainting + Occlusion detector   & \cellcolor{yellow!15}0,9201 &  \cellcolor{yellow!15}24,8619 & \cellcolor{yellow!15}0,0905 \\ \hline
Both + Inpainting + Ocl. + RGB tuning & \cellcolor{green!15}0,9203 &  \cellcolor{green!15}24,9758   & \cellcolor{green!15}0,0901                    
\end{tabular}

\caption{\label{table:ablation} \textbf{Quantitative ablation study.} Metrics were measured for images of 15 people with diverse poses and view points.}
\vspace{-1em}
\end{table}

\subsection{Quantitative results}

%The quantitative comparison is shown in Table~\ref{table:metrics}. We compare our approach with various methods for generating an avatar from a single image, including those requiring additional rigging steps for animation. For clarity, the table is divided into three sections. PIFu~\cite{saito2019pifu} and PHORHUM~\cite{alldieck2022photorealistic} restore the 3D mesh of a person in a pose shown in the image. This imposes strong restrictions on the pose of a person in the input image if one wants to animate it. ARCH~\cite{huang2020arch} and ARCH++~\cite{he2021arch++} restore the 3D mesh in canonical space, which is easier to rig and animate. StylePeople~\cite{grigorev2021stylepeople} and DINAR (ours) are based on a parametric human model and therefore are the easiest to animate and do not suffer from rigging imperfections.

To numerically evaluate our avatars, we report several metrics (Table~\ref{table:metrics}). For clarity, the table is divided into three sections: PIFu~\cite{saito2019pifu}, PHORHUM~\cite{alldieck2022photorealistic} - require additional rigging; ARCH~\cite{huang2020arch}, ARCH++~\cite{he2021arch++} - restore geometry in the canonical space; StylePeople~\cite{grigorev2021stylepeople}, DINAR (ours) - utilize a parametric human model and therefore can be animated without additional procedures. We measured reference-based metrics in the front view avatars for the SnapshotPeople public benchmark, namely: Multi-scale structural similarity (MS\nobreakdash-SSIM $\uparrow$), Peak signal-to-noise ratio (PSNR $\uparrow$), Learned Perceptual Image Patch Similarity (LPIPS $\downarrow$)~\cite{zhang2018unreasonable}. We found that our method works on par with non-rigged methods.

To evaluate the quality of the back view (and thus the ability to generalize to new views), we report Kernel Inception distance (KID $\downarrow$)~\cite{binkowski2018demystifying} measurements. This metric allows one to evaluate generated images quality and is more suitable for small amounts of data than FID~\cite{heusel2017gans}. Our approach results in the highest KID value compared to other methods. To assess identity preservation we measured re-identification score (ReID $\downarrow$) based on FlipReID~\cite{Ni2021FlipReIDCT} model for human re-identification. Our method shows the best results in person's identity preserving between front and back views. To additionally validate the quality of the textures and to measure structural similarity in cases with unaligned ground truth images we measured Deep Image Structure and Texture Similarity (DISTS~$\downarrow$)~\cite{ding2020image}. We provide measurements for front view and back view in the table. Our method produce the most naturally looking avatars from both views.

\subsection{Qualitative Results}

\begin{figure*}
  \centering
  \includegraphics[width=0.8\linewidth]{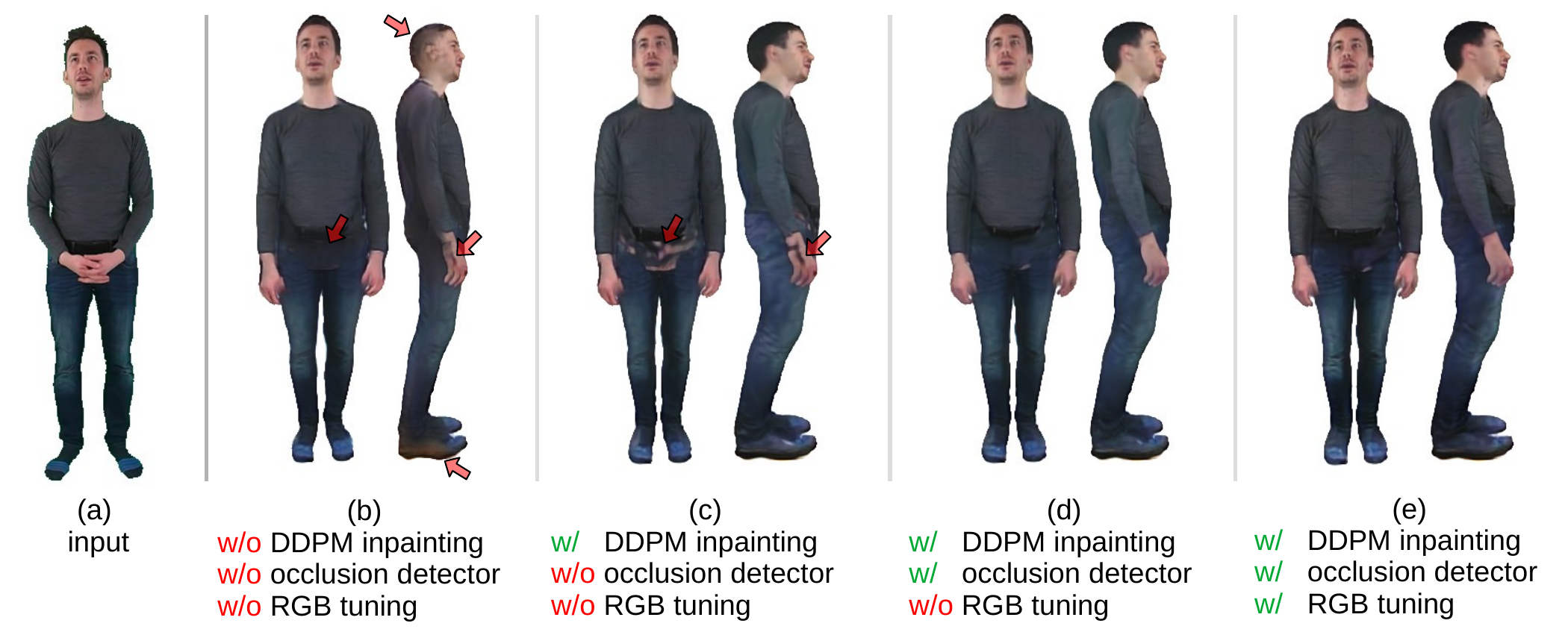}
   \caption{\textbf{Ablation study.} b) Avatars without DDPM inpainting are prone to artifacts and lack of realism. c) Disabling the occlusions detector leads to artifacts due to sampling errors at the edge of the overlapped areas. d) Disabling RGB finetuning results in less high-frequency detail and worse color matching. e) The best result is achieved by using all the steps of the pipeline.}
   \label{fig:Ablation}
\end{figure*}

\begin{figure}
  \centering
  \includegraphics[width=0.8\linewidth]{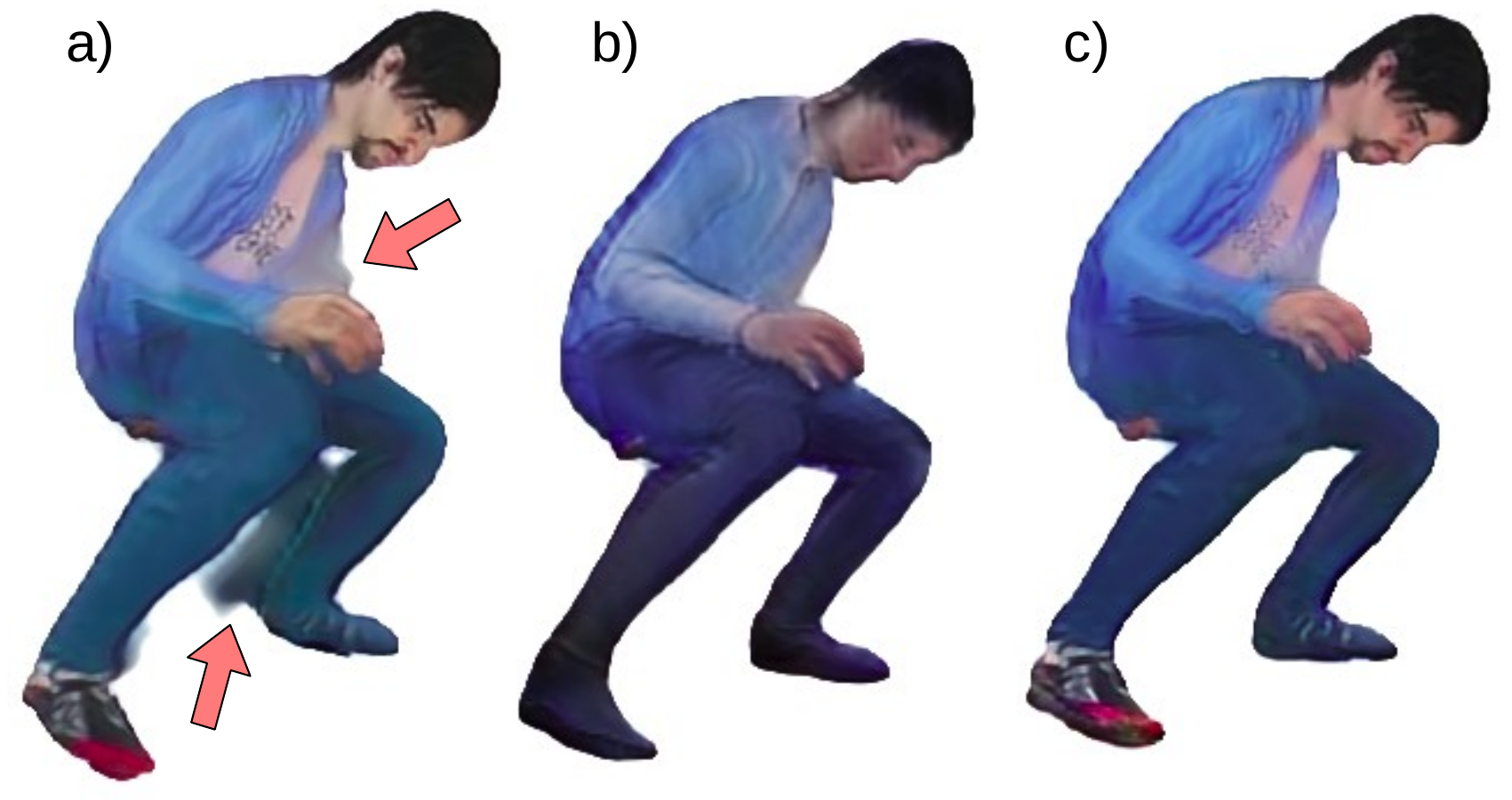}
   \caption{\textbf{Texture ablation study.} a) RGB texture; b) Neural texture; c) Both textures.}
   \label{fig:SG2_ablation}
\end{figure}

Metric value does not always correlate well with human perception. In Fig~\ref{fig:Snapshot}, we show the qualitative results of our method in comparison with other one-shot avatars methods: PIFu, PHORHUM, ARCH, ARCH++ and StylePeople. The figure shows both front and back views of the avatars. Additional comparisons are shown in Supplementary materials, including results on the THuman2.0 dataset \cite{tao2021function4d}.

Overall, our method realistically reconstructs the texture of clothing fabrics on the back (\eg pleats on pants), which boosts the realism of the renders. Using the whole information from the given image allows us not to copy unwanted patterns from front to back (as is commonly done by the pixel-aligned methods while recovering the texture for the back). Using sampled RGB texture as an addition to a neural texture allows us to achieve photo-realistic facial details and preserve high frequency details. We note that PIFu accurately reproduces the color of the avatar and restores the geometry well. However, it does not preserve high-frequency details, which is why avatars suffer from a lack of photo-realism. PHORHUM generates highly photo-realistic avatars but often suffers from color shifts. Another methodological shortcoming of this approach is the absence of a human body prior. Therefore, the model can be over-fitted on training dataset's human poses, which may lead to incorrect work with unseen poses. Avatars generated by ARCH contain strong color artifacts and suffer from geometry restoration errors. ARCH++ significantly improves geometry and color quality for the frontal view, but the back view still suffers from color shift and artifacts. StylePeople is based on a parametric human model and can be easily animated without the use of third party methods or additional rigging. However, the coverage of the latent space of their model is limited, which leads to overfitting and poor generalization to unseen people, when performing inference based on a single view.

\section{Ablation study}

We quantitatively (Table \ref{table:ablation}) and qualitatively (Fig. \ref{fig:Ablation}, \ref{fig:SG2_ablation}) ablate texture choice and the proposed method steps. For this, we used 15 images from AzurePeople~\cite{bashirov2021real} with diverse poses and view points. Eight people's input images from the front view point, and seven from the back view point. 

We have studied the efficiency of using RGB and neural channels in the texture for the one-shot avatar task. We found that using both kinds of channels results in better metrics than using them separately. We also studied the impact of using a DDPM inpainting model (Section \ref{sec:Inpainting}) on the final metrics. In scenarios without a model for \textit{Inpainting}, the renderer takes over this function. The \textit{Occlusion detection} step is used to remove RGB texture sampling errors caused by SMPL-X inaccuracy and is described in section \ref{sec:Oclusion}. The final step in the ablation study is \textit{RGB channels fine-tuning} on the input image for a few steps (Section \ref{sec:Inference}). 

The best metrics are achieved using all the steps of the proposed approach. We also provide a visual comparison of the various steps of the ablation study in the Figure \ref{fig:Ablation}.

Figure~\ref{fig:SG2_ablation}~(a) illustrates how the renderer is unable to obtain information about the out-of-mesh details while using only the RGB texture. This leads to artifacts around the avatar in new poses. In turn, the use of only neural channels leads to the loss of high-frequency information (Fig. \ref{fig:SG2_ablation} (b)). Using both types of textures allows us to pass information about out-of-mesh elements to the renderer and preserve high-frequency details (Fig. \ref{fig:SG2_ablation} (c)).

%-------------------------------------------------------------------------

\section{Conclusion}

We have presented a new approach for modeling human avatars based on neural textures that combines the RGB and the latent components. The RGB components are used to preserve the high frequency details, while the neural components add hair and clothing to the base SMPL-X mesh. Using the parametric SMPL-X model as a basis makes it easy to animate the resulting avatar. Our method restores missing texture parts using an adapted diffusion framework for inpainting such textures. Our method thus creates rigged avatars, while also improving the rendering quality of the unseen body parts when compared to modern non-rigged human model reconstruction methods.

{\small
\bibliographystyle{ieee_fullname}
\bibliography{egbib}
}

% \iffalse

\clearpage
\setcounter{section}{0}
\renewcommand{\thesection}{\Alph{section}}

\section{RGB sampling algorithm}

In our approach, we use the combination of RGB and neural texture. To sample the RGB texture, we use the following algorithm:

\begin{algorithm}
\caption{RGB texture sampling algorithm}\label{alg:Sampling}
\begin{algorithmic}
\Require $RGB(\textrm{size} \times \textrm{size} \times 3)$
\Require $UV(\textrm{size} \times \textrm{size} \times 2)$
\State \textcolor{teal}{\# Initialize texture with zeros}
\State $T \gets zeros(\textrm{texture\_size} \times \textrm{texture\_size} \times 3)$ 
\State $C \gets zeros(\textrm{texture\_size} \times \textrm{texture\_size})$
\State
\State \textcolor{teal}{\# Fill texels with mean value of neighbors} 
\For{$ \forall x, y \in [0 .. \textrm{size}]$}
   \State $(i,j) \gets UV[x, y]$
   \For{$\forall k, m \in [-1, 0, 1]$}
       \State $T[i + k, j + m] \mathrel{+}= RGB[x, y]$
       \State $C[i + k, j + m] \mathrel{+}= 1$
   \EndFor
\EndFor
\State $T = T / C$
\State
\State \textcolor{teal}{\# Fill exact values in texels we don't need to inpaint}
\For{$ \forall x, y \in [0 .. \textrm{size}]$}
   \State $(i,j) \gets UV[x, y]$
   \State $T[i, j] \gets RGB[x, y]$
\EndFor
\end{algorithmic}
\end{algorithm}

A simple filling with an average value is needed in order to remove the gaps that appear on the texture due to the discreteness of sampling grid. The described algorithm allows us to fill them taking into account the color of neighboring texels.

\begin{figure*}
  \centering
  \includegraphics[width=1.0\linewidth]{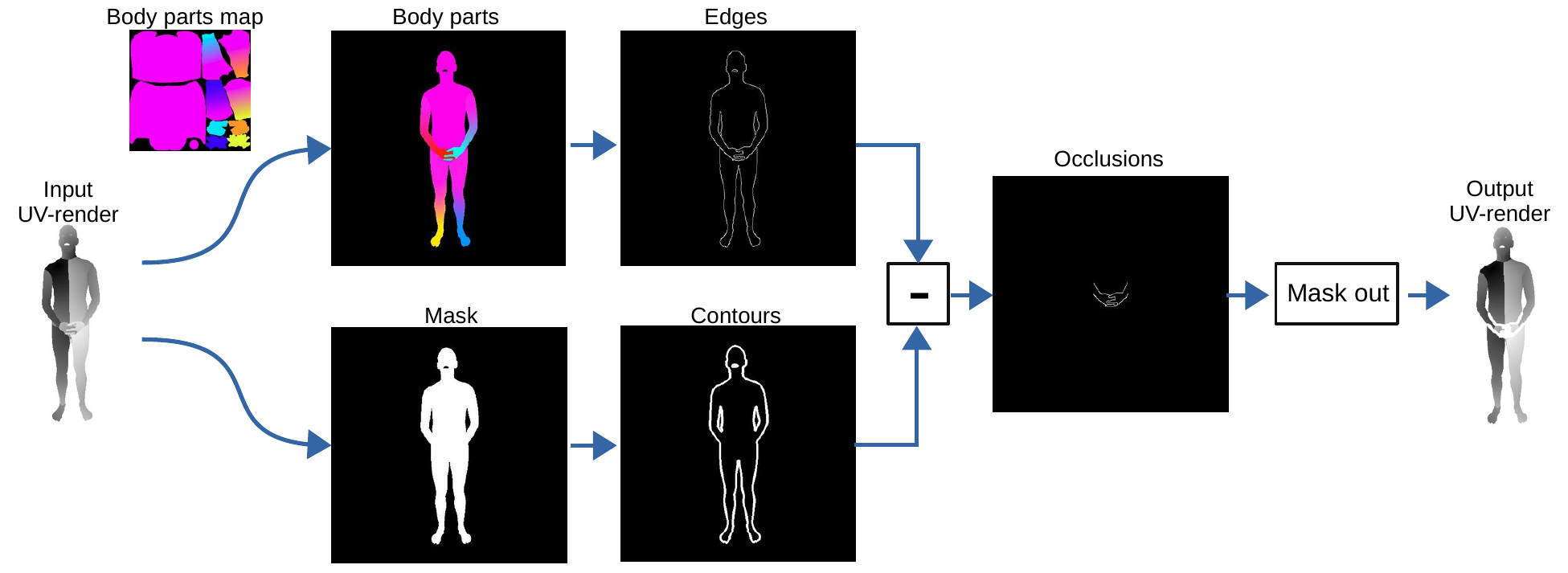}
   \caption{\textbf{Occlusions detection.} We use the body parts map as a texture to detect self-occluded areas on the avatar. From UV-render we will get rendered body parts. We get the outer silhouette of the avatar by binarization. We detect the outlines of the avatar with the edge detector. Based on the difference of contours and edges, we determine the outer contour of the occlusion area and remove it from the UV-render.}
   \label{fig:Occlusions}
   \vspace{-1em}
\end{figure*}

In order to avoid sampling errors caused by the inaccuracy of the SMPL-X fitting, we used the occlusion detector described in Section \ref{sec:Oclusion}. We have shown a more thorough diagram of this stage in Figure \ref{fig:Occlusions}.

\section{RGB texture refinement} \label{sec:RGBrefine}

We have found it beneficial to perform several optimization steps (namely 64) of RGB texture to enhance high frequency details (Fig. \ref{fig:Ablation} (e)) in the inference stage. To achieve this, we use gradients from the neural renderer derived by comparing the rendering result with the input image. Gradients are applied to texels with weights that correspond to the angles between the normal vectors and the camera direction (Fig. \ref{fig:Merge}). This makes sure that only texels that can be seen in the input image are optimized with prioritization of more frontal ones. We employed $L2$ and $LPIPS$ losses to encourage color matching, and $Adversarial$ loss with regularization analogous to the \ref{eq:losses} equation to amplify detalization. 

We also apply a linear adjustment to the RGB channels of the VQGAN decoding output to improve color matching between front and back views after the inpainting stage: 
\begin{equation}
T_\textrm{rgb} = T_\textrm{rgb} \alpha + \beta.
\end{equation}

In this case, all texels share the trainable parameters alpha and beta. We optimize them with renderer's gradients derived by the pixels visible in the input image. As a result, the RGB channels of the neural texture at the VQGAN output strengthen color matching with the sampled RGB texture.
This helps us to minimize the seam after combining textures (Fig. \ref{fig:Ablation} (e)). 

\section{Architecture details}

\begin{figure*}
  \centering
  \begin{subfigure}{0.35\linewidth}
    \centering
    \includegraphics[width=1.0\linewidth]{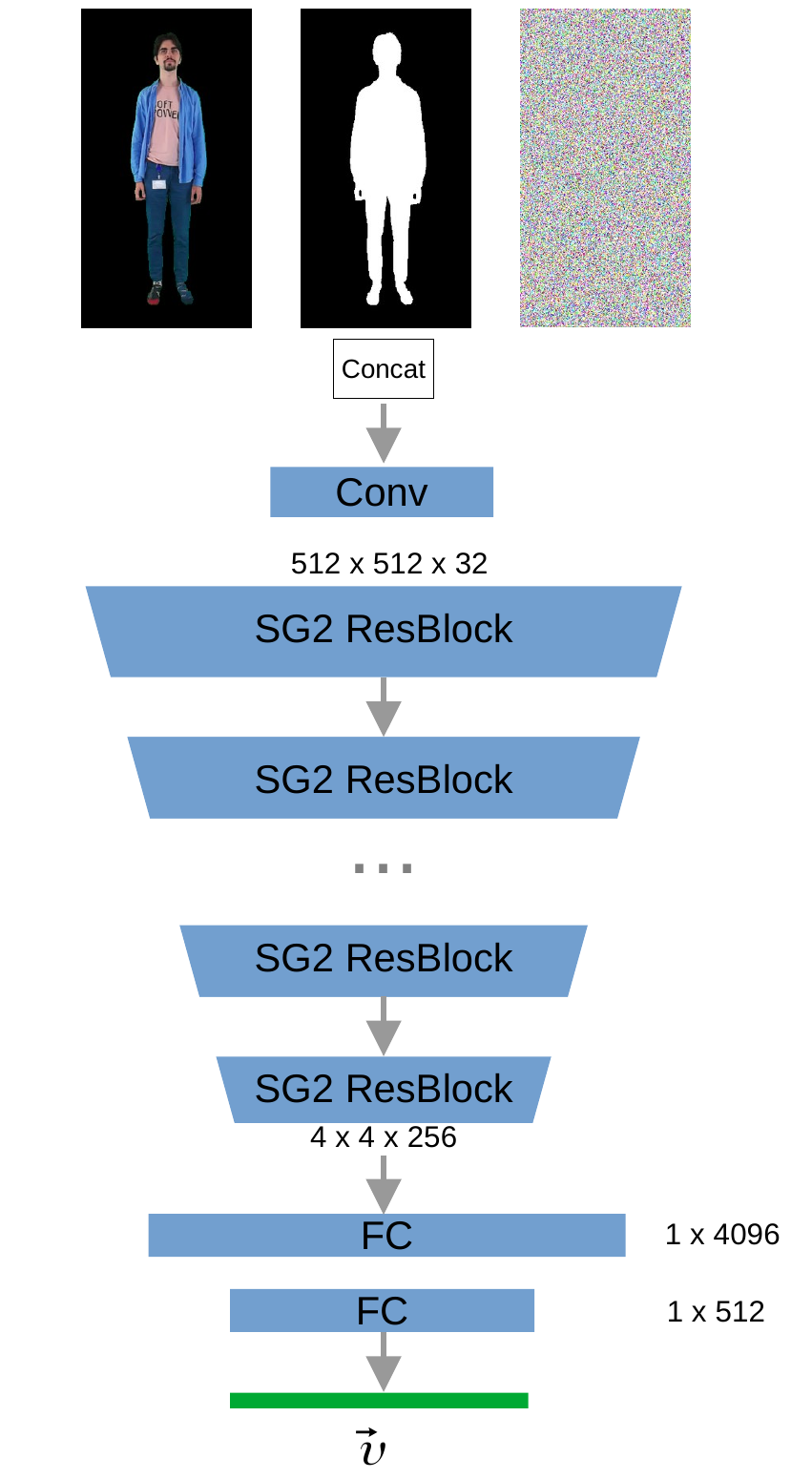}
     \caption{\textbf{Encoder architecture}}
     \label{fig:Arch-a}
  \end{subfigure}
  \begin{subfigure}{0.35\linewidth}
    \centering
    \includegraphics[width=0.76\linewidth]{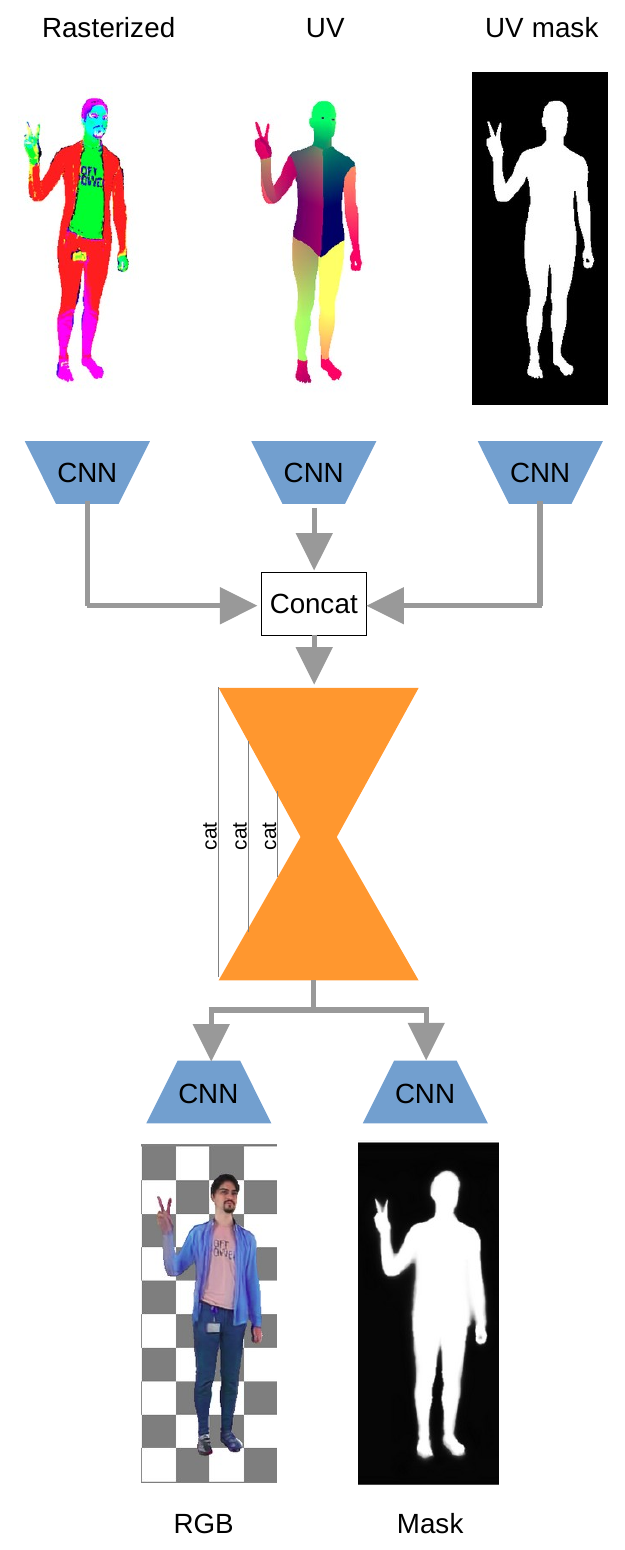}
     \caption{\textbf{Renderer architecture}}
     \label{fig:Arch-b}
  \end{subfigure}
  \caption{\textbf{Encoder and renderer architecture.} The encoder architecture is a modified architecture of the StyleGAN2 discriminator. We changed the head to get a vector of length 512. The renderer has a U-Net architecture that predicts the RGB of an avatar from a rasterized model and UV render.}
  \label{fig:Arch}
\end{figure*}

\textbf{Encoder network.}
As an encoder network (Fig. \ref{fig:Arch-a}), we have adapted the StyleGAN2 discriminator architecture with a few changes. Namely, three images are fed to the network input: RGB, segmentation mask, and single-channel noise. Noise is introduced to provide additional freedom to the generative model when training the GAN. The efficiency of using noise in generative neural networks has been demonstrated by the authors of StyleGAN.

The images received at the input are concatenated by channels and passed through a feature extractor with an architecture equivalent to the StyleGAN discriminator consisting of ResNet blocks. We modified the model head so that it outputs a vector of length 512. This vector is then used as the input of the StyleGAN2 generator and the proposed encoder is trained end-to-end with the generator and the renderer.

Our model is trained on RGB images at the $512 \times 512$ resolution. For each $3 \times 512 \times 512$ input image, we generate a $21 \times 256 \times 256$ neural texture. In the texture, the first 16 channels are generated by the network $G$, the next three channels are RGB channels and the remaining two channels are the sampling and the inpainting masks.

\textbf{Renderer network.}
Here we describe the $\theta$ renderer (Fig. \ref{fig:Arch-b}). The resulting texture is applied to the SMPL-X model and rasterized. The rasterized image has a size of $21 \times 512 \times 512$ and is fed to a neural renderer. The renderer takes three images as input: a rasterized SMPL-X model with a neural texture, a UV render, and a UV mask. Each input image is passed through a convolutional network consisting of two convolutions with LeakyReLU activation and BatchNorm layers. Output features are concatenated and fed into a U-Net consisting of ResNet blocks. U-Net has 3 levels connected by feature concatenation. The U-Net output is passed through two additional convolutional networks to predict the RGB image of the avatar and its mask.

\section{Robustness of the method}

\begin{figure*}
  \centering
  \includegraphics[width=1.0\linewidth]{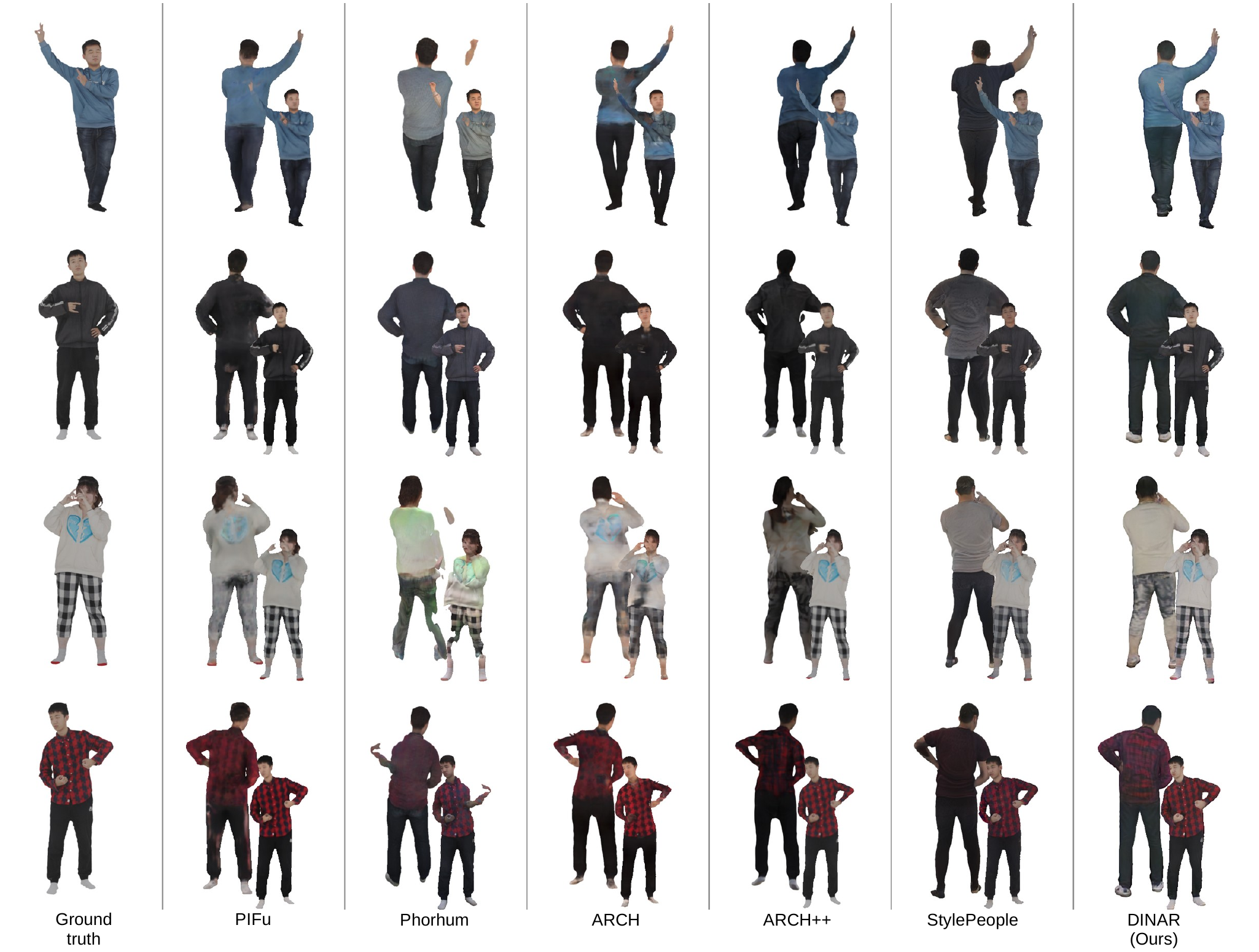}
   \caption{\textbf{Results on the THuman2.0 dataset.} We compared our method with existing approaches on an additional dataset. Similar to the main article, our method shows the most convincing results for the new dataset.}
   \label{fig:THuman}
\end{figure*}

In order to assess the robustness of our method, we compared the methods qualitatively (Fig. \ref{fig:THuman}) and quantitatively (Table \ref{table:metrics_thuman}) on the additionally dataset: THuman2.0\cite{tao2021function4d}. This dataset contains 3D scans of people in diverse clothes and complex poses. For the test, we selected 25 random people and used the front-view renders as input for the one-shot methods. Our method obtains the most convincing front and back views and is resistant to complex poses and various datasets (Fig. \ref{fig:THuman}). This is also confirmed by objective metrics (Table \ref{table:metrics_thuman}). Our method allows us to get the best metrics for the novel view on this additional dataset.

\begin{figure*}
  \centering
  \includegraphics[width=0.6\linewidth]{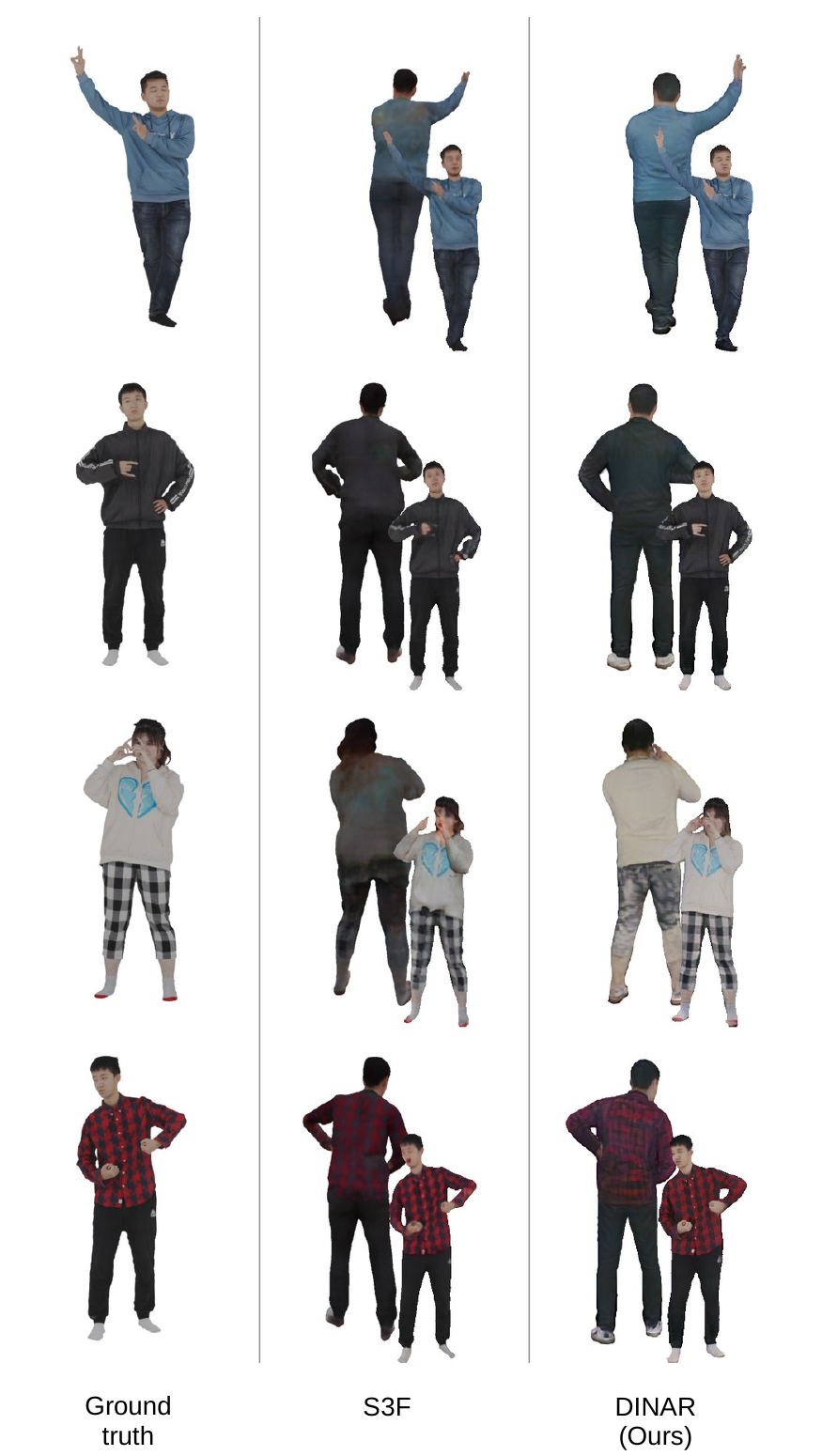}
   \caption{\textbf{Comparison with S3F on the THuman2.0 dataset.} We compared our approach with the most recent one-shot approach: Structured 3D Features.}
   \label{fig:S3F}
\end{figure*}

We also used THuman2.0 to compare with the recent S3F~\cite{corona2023structured} method for generating one-shot avatars (Fig. \ref{fig:S3F}). Our approach better preserves high-frequency detail in the front view and produces fewer artifacts in the back view. However, their method better restores a regular pattern on the back and allows model relightning.

\begin{table*}[]
\centering
\begin{tabular}{c|cccc|ccc}
                     & \multicolumn{4}{c|}{\textbf{Same view}} & \multicolumn{3}{c}{\textbf{Novel view}} \\
Method               & MS-SSIM $\uparrow$ & PSNR $\uparrow$ & LPIPS $\downarrow$ & DISTS $\downarrow$ & DISTS $\downarrow$ & ReID $\downarrow$\  & KID $\downarrow$ \\ \hline
PIFu                 & \cellcolor{green!15}0,9893  & \cellcolor{green!15}28,2686   & \cellcolor{green!15}0,0474  & \cellcolor{yellow!15}0,0912 &     0,2213       &      0.11608           &    0,0924          \\
Phorhum              &    0,9566                   &  23,8915                      &        0,0521               &    0,1249                   & 0,1835 &      0.12516           &   0,0389    \\ \hline
ARCH                 &    0,9372                   &  21,7770                      &        0,0645               &    0,1617                   &     0,2082       &     0.12193            &     0,1065         \\
ARCH++               &    0,9577                   &  22,8318                      &        0,0562               &    0,1067                   &     \cellcolor{yellow!15}0,1806       & \cellcolor{yellow!15}0.10108   &     0,0408         \\ \hline
S3F              &    0,9706                   &  25,6459                      &        \cellcolor{yellow!15}0,0497               &    0,1152                   & 0,1928 &      0.11991           &  0,1108    \\
StylePeople          &    \cellcolor{yellow!15}0,9765                   &  \cellcolor{yellow!15}25,8120                      &        0,0588               &    \cellcolor{green!15}0,0830                   &     0,1828       &     0.14212            &     \cellcolor{yellow!15}0,0347         \\
\textbf{DINAR (Ours)} & 0,9600 &  22,8671 & 0,0568 & 0,0975  &  \cellcolor{green!15}0,1607    &  \cellcolor{green!15}0.09999  & \cellcolor{green!15}0,0250                          
\end{tabular}

\caption{\label{table:metrics_thuman} \textbf{Metrics comparison on the THuman2.0 dataset.} To demonstrate the robustness of our approach, we evaluated the metrics on a second dataset. The table is compiled similarly to Table 1 from the main paper.}
%\vspace{-1em}
\end{table*}

\section{Additional results}

\begin{figure*}
  \centering
  \includegraphics[width=1.0\linewidth]{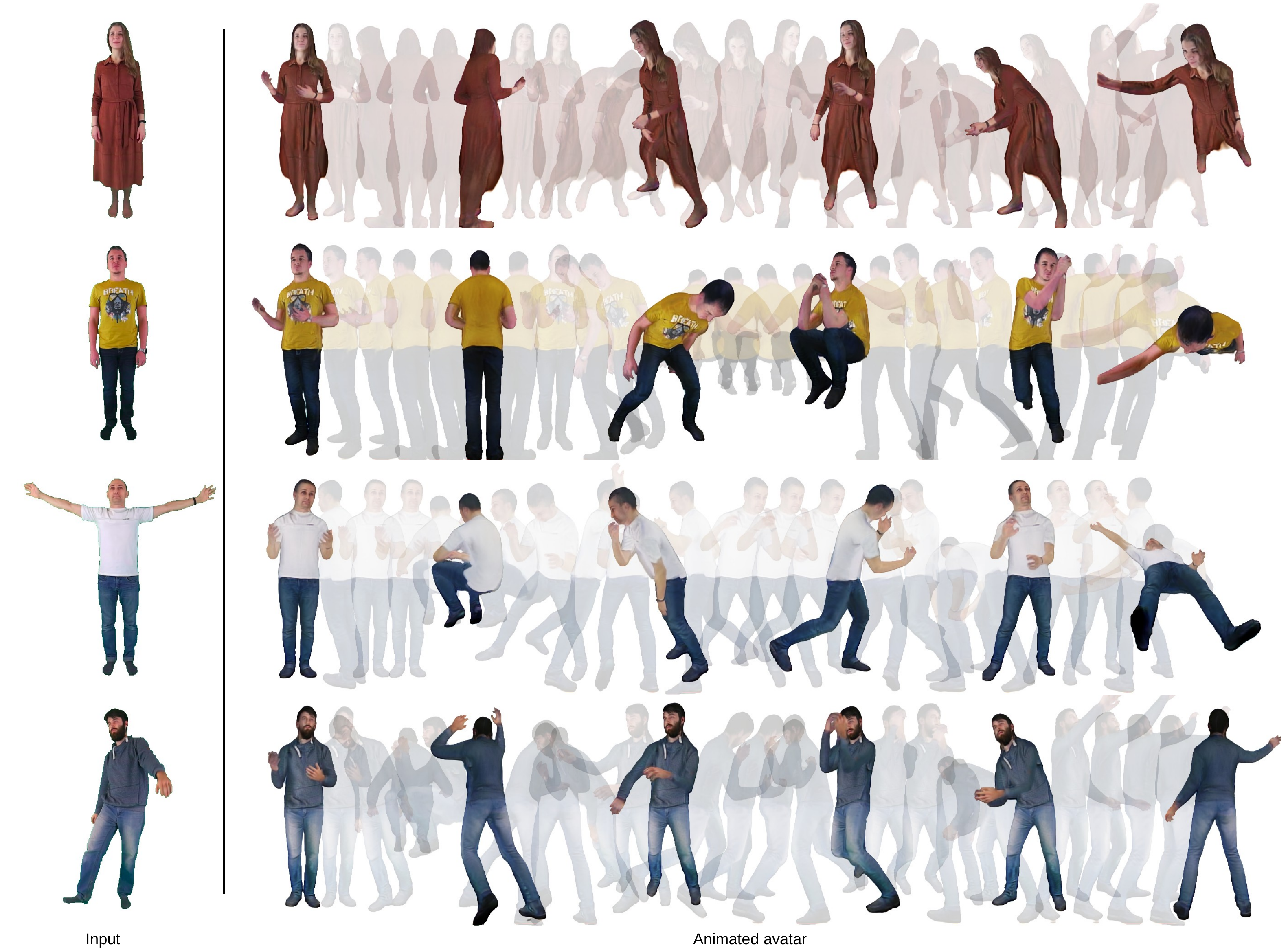}
   \caption{\textbf{More avatar animation examples.} We present more examples of avatar animations, including those obtained from more complex poses. The top two rows demonstrate how the approach works with people in loose clothes and clothes with highly detailed prints.}
   \label{fig:Additional}
\end{figure*}

We present additional results of our approach on diverse data. On Fig. \ref{fig:Additional} we show results for input images containing different people. The top row shows an additional example of processing of a person in loose clothing. The next row demonstrates the high-fidelity rendering of an avatar wearing a T-shirt with a complex high-frequency print. The bottom two rows demonstrate the accuracy of avatar reconstruction from images of people in unusual poses. Also, the frames of the animation sequence show the avatars from more varied viewpoints (\eg top and bottom). Invariance to the human pose is achieved through the use of a neural texture framework with a parametric model. All avatar processing, such as restoring the back, is done in canonical texture space. 

\begin{figure*}
  \centering
  \includegraphics[width=1.0\linewidth]{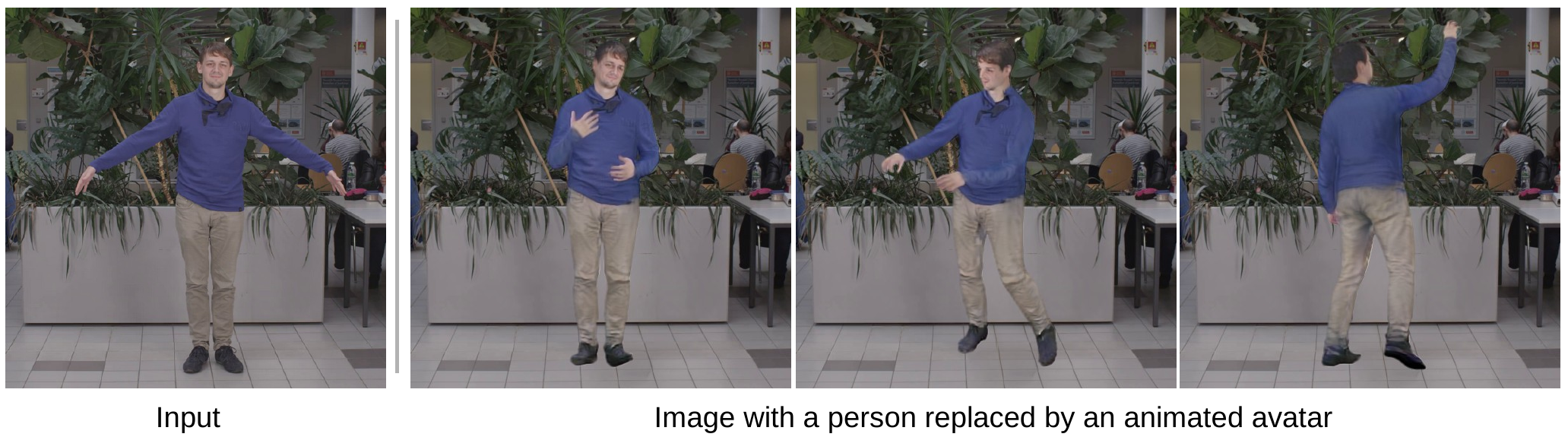}
   \caption{\textbf{Making photos come alive.} An additional use case of our approach is to replace the person in the photo with their animated avatar. By doing this, we can achieve the effect of an animated photo.}
   \label{fig:Alife}
\end{figure*}

On Fig. \ref{fig:Alife} we demonstrate an additional use case for our one-shot approach. We used neural network inpainting to remove the person from the original image and replace it with an animated avatar. In this way we can create the effect of a photo that has come to life.

One of the limitations of the current approach is the handling of tissue deformations in the input image. Our method does not modify the textures depending on the pose, which can make the fabric look unrealistic when changing the pose. Another limitation is the insufficient sharpness of the edges of loose clothing. Even though dresses are rendered correctly by our method on most frames, the edges of the dress look unrealistic. In our future research, we would like to focus on addressing these two shortcomings.

%\fi

\end{document}